\documentclass[11pt,a4paper]{article}

\usepackage[pdftex]{graphicx}
\usepackage{subcaption}

\usepackage{url}

\usepackage[font=small]{caption}
\usepackage[table]{xcolor}

\usepackage{amssymb,amsmath}
\usepackage{mathrsfs}
\usepackage{txfonts}
\usepackage{url}
\usepackage{algorithmic}

\title{{\bf SmileyNet} - Towards the Prediction of the Lottery \\ by Reading Tea Leaves with AI}
\author{Andreas Birk\\
Constructor University Bremen\\
School of Computer Science and Engineering\\
Campus Ring 1, 28759 Bremen \\
abirk@constructor.university
}
\date{\today}

\begin{document}
\twocolumn

\DeclareGraphicsExtensions{.jpg,.pdf,.mps,.png}

\maketitle

\begin{abstract}
We introduce SmileyNet, a novel neural network with psychic abilities. It is inspired by the fact that a positive mood can lead to improved cognitive capabilities including classification tasks. The network is hence presented in a first phase with smileys and an encouraging loss function is defined to bias it into a good mood. SmileyNet is then used to forecast the flipping of a coin based on an established method of Tasseology, namely by reading tea leaves. Training and testing in this second phase are done with a high-fidelity simulation based on real-world pixels sampled from a professional tea-reading cup. SmileyNet has an amazing accuracy of 72\% to correctly predict the flip of a coin. Resnet-34, respectively YOLOv5 achieve only 49\%, respectively 53\%. It is then shown how multiple SmileyNets can be combined to win the lottery.     
\end{abstract}

\section{Introduction}
Modern AI is one of the greatest if not {\em the} greatest scientific breakthrough of all times. The publication of the structure of DNA by Watson and Crick in Nature \cite{WatsonCrick-DNA-Nature1953} has only achieved slightly more than 19,000 citations by summer 2024 according to GoogleScholar. And it has taken 71 years. AI \cite{DeepLearning-Hinton-Nature15,DeepLearning-Book-MITpress16} has led to many important papers, which received many more citations within a much shorter time-frame; see, e.g., Tab.~\ref{tab:cites} for a few examples based on citations according to GoogleScholar.

\begin{table*}[!htb]
	\centering
	\begin{tabular}{|l|c|c|c|c|c|c|c|c|c|c|c|}\hline
	 & {\bf DNA} & \multicolumn{10}{c|}{{\bf AI}} \\ \hline
  publication & \cite{WatsonCrick-DNA-Nature1953}	& \cite{DeepLearning-ResNet-CVPR16}	& \cite{Adam-StochasticOptimization-arXiv2014}	& \cite{DeepLearning-AlexNet-NIPS12}	& \cite{Breimann-RandomForests-ML01}	& \cite{DeepCNN-LargeScale-ImageRecognition-2014}	& \cite{DeepLearning-Hinton-Nature15}	& \cite{GAN-NIPS2014}	& \cite{FasterR-CNN-NIPS2015} & \cite{ImageNet-CVPR09} & \cite{DeepLearning-GoogLeNet-CVPR15}\\
  year & 1953	& 2016	& 2014	& 2012	& 2001	& 2014	& 2015	& 2014	& 2015 & 2009	& 2015\\
  citations (x 1K) & 19	& 232	& 189	& 159	& 133	& 127	& 81	& 79	& 76 & 71	& 61\\
  citations/year (x 1K)& 0.27 	& 29	& 19	& 13	& 6	& 13	& 9	& 8	& 8 & 5	& 7 \\ \hline
  \end{tabular}
\caption{The scientific relevance of AI compared to the discovery of DNA. \label{tab:cites}}
\end{table*}

Given the relevance of AI for our modern world, it is surprising that there are areas of high economic relevance where it has not been applied, yet. We explore here - up to our knowledge for the very first time - the potential of AI for the psychic services industry, which is a multi-billion dollar business. This industry is estimated to have reached a revenue of well above \$2 billion in the USA in 2023. And it experiences a steady further growth. This holds especially for psychic online services that managed an exponential growth in recent years according to market-reports.       

We suspect that AI may already be used in some real-world psychic services, especially in the online branches. But the methods and the results are not academically published up to our knowledge - likely due to the large economic potential they possess. While this is somewhat understandable, we believe in Open Science \cite{UNESCO-RecommendationOpenScience2021} and we share our insights due to the large benefits Psychic AI can provide to humanity.

A first medium-term goal is to win the lottery to further fund research in the novel field of Psychic AI. A well-established method of Tasseology (also known as Tasseomancy or Tasseography) \cite{Tasseology-2008} is used to this end, namely reading tea leaves \cite{Tasseology-TeaLeaves-2016,Tasseology-TeaLeaves-2014}. In the spirit of divide and conquer, the simpler problem of predicting the outcome of flipping a coin is tackled first. We then show how this can be used to predict the numbers for winning the lottery jackpot. 

As an important basic research result, it is proven in this paper, up to our knowledge for the very first time, that a neural network can possess psychic powers. Concretely, it is shown that the novel SmileyNet can predict the flipping of a coin with a precision well above chance. 


\section{SmileyNet}
\label{sec:SmileyNet}
SmileyNet is a novel bio-inspired network. It is motivated by the fact that cognitive processes can be positively influenced by a positive mood 
\cite{PositiveMood-CreativeProblemSolving-Sage2015,PositiveMood-CognitiveControl-2011}.
This includes especially also classification tasks \cite{PositiveMood-Cognition-Classification-2010}. 

\begin{figure}[htbp]
\begin{center}
\includegraphics[width=.5\linewidth]{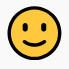}
\caption{The smiley in form of the Unicode emoticon "U+1F642" that is used to bias the network into a positive mood. \label{fig:smiley}}
\end{center}
\end{figure}

Furthermore, it is well-known that visual stimuli are well suited to elicit emotions \cite{ImagesMoodSurvey2016,ImagesMoodSurvey2008}. This includes in particular a smiling face \cite{MoodSmilingFace2019}. We hence first present the network with smileys as input to bias it into a good mood. More precisely, the Unicode emoticon "U+1F642" for the classical smiley, officially known as "Slightly Smiling Face" is used to this end (Fig.\ref{fig:smiley}).

But stress is a negative factor for cognition \cite{CognitionStressNegative2024,CognitionStressNegative2023}. We do not want to loose the positive effects of the good mood by putting the network under stress in the smiley training phase. Hence, it gets positive feedback \cite{StressCognitionFeedback2020} that it is doing fine, i.e., that it is doing nothing wrong.

More precisely, the training in the {\em smileyfication} phase is done as follows. A quadratic loss function ${\cal L}_{smile}()$ is used. To avoid putting the network under stress and to give a further positive endorsement, the expected output $\hat{y}$ is set to be the actual output $y$ whenever a smiley is presented to the net, i.e., $\hat{y} = f_I(y)$ where $f_I()$ is the identity function. So,

\[{\cal L}_{smile}()= \frac{1}{2} (y - \hat{y})^2 \text{ with } \hat{y} = f_I(y)\]

This method of smileyfication can be applied to any network ${\cal N}_{old}$ to turn it into a novel one ${\cal N}_{new}$, which is expected to perform better than the previous ${\cal N}_{old}$. 

Often, there may only be small improvements and it can happen that multiple trials are needed until some improvement is seen for the first time, which should then be immediately documented. Also, it can be useful to take additional factors into account. For example, the effects of smileyfication may only be seen on a sunny day as, e.g., the weather has an influence on the mood \cite{MoodWeather2008,MoodWeather1984}.  

\section{Related Work}
\label{sec:sota}
YOLOv5 \cite{YOLOv5-2021} and Resnet34 \cite{DeepLearning-ResNet-CVPR16} are well-known networks for image classification. They are used for comparison to SmileyNet in Sec.~\ref{sec:results}. There are also many other networks. But as YOLOv5 and Resnet32 are very popular, we concentrate here on those two. 

\begin{figure}[htbp]
\begin{center}
\includegraphics[width=\linewidth]{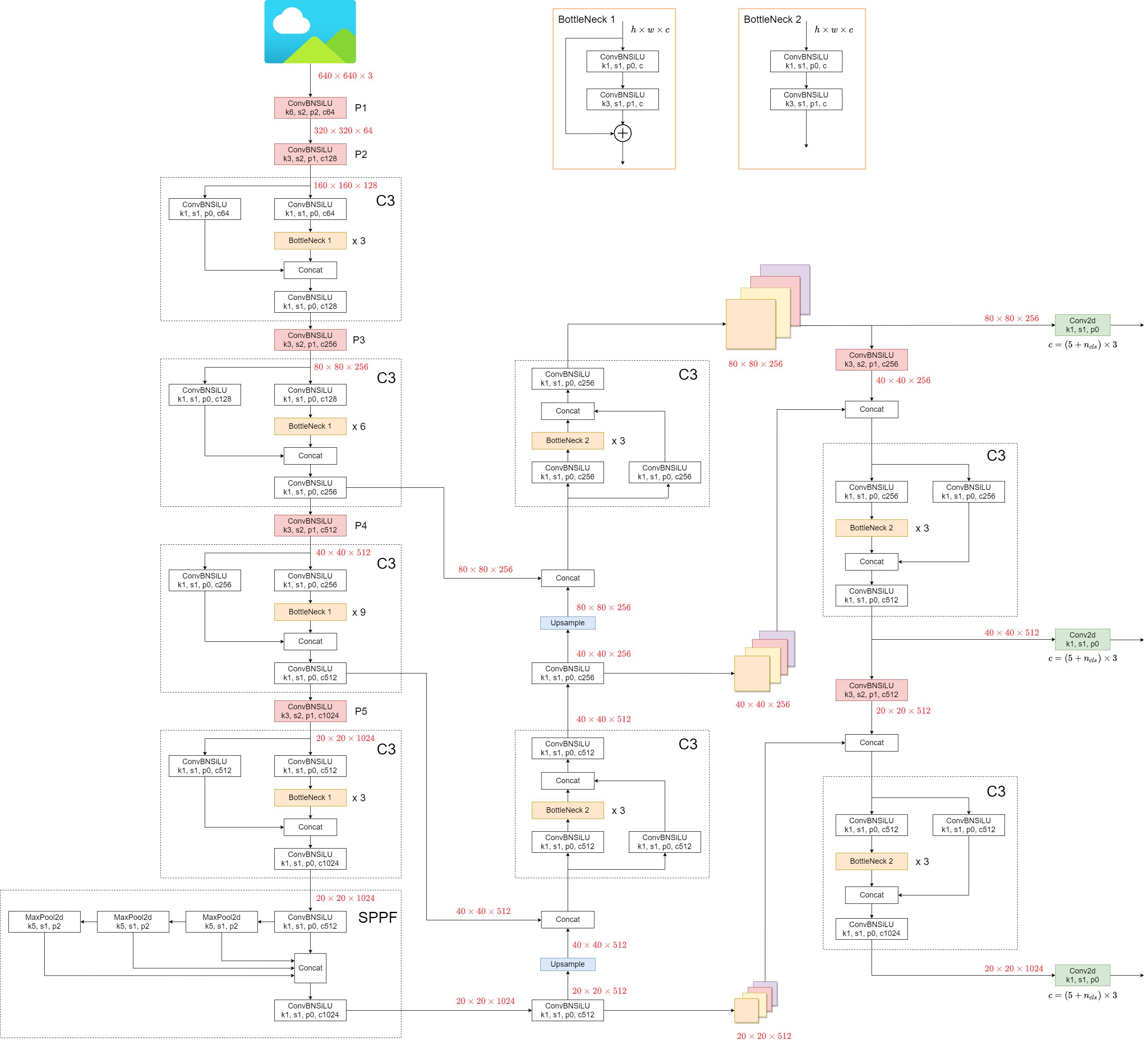}
\caption{YOLOv5 \cite{YOLOv5-architecture-2021} \label{fig:YOLOv5}}
\end{center}
\end{figure}

\begin{figure}[htbp]
\begin{center}
\includegraphics[width=\linewidth]{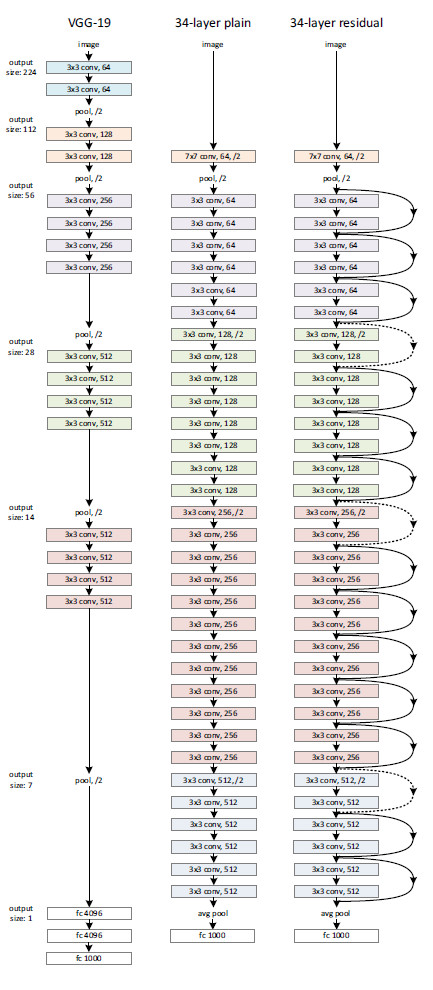}
\caption{Resnet \cite{DeepLearning-ResNet-CVPR16} \label{fig:Resnet}}
\end{center}
\end{figure}

YOLOv5 is the fifth iteration of the revolutionary "You Only Look Once" object detection model. It is designed to deliver high-speed, high-accuracy results in real-time. Built on PyTorch, this powerful deep learning framework has garnered immense popularity for its versatility, ease of use, and high performance. Fig.~\ref{fig:YOLOv5} shows how it works.

ResNet-34 is a convolutional neural network in PyTorch. It is has 34 layers. It addresses the problem of vanishing gradient. This is done with the identity shortcut connection that skips one or more layers. 
Fig.~\ref{fig:Resnet} shows how it works.

\section{High-Fidelity Simulation}
\label{sec:sim}
The constellation of tea leaves and the predicted event are linked by cosmic energy. We use this fact to generate high-fidelity training data. This is not easy in the context of psychic powers and it may be one reason why AI has not been used - up to our knowledge - in this economically highly relevant business up to now.

The core idea of our approach for training data is to use the reverse direction of the link. The coin is flipped in a deterministic way, i.e., it is so to say turned by hand in rounds $r$ from one side to the other. The related constellation of tea-leaves, i.e., the image $I_r$ is then generated afterwards. 

In the software implementation, "heads" is presented with a "1" and "tails" with a "0" for the sake of convenience and wlog. Also wlog, we start with "tails" facing up as initialization of the flipping process. In the very first round $r=1$, the coin is turned and it hence shows "heads". The coin is then turned again and it now shows "tails" in round $r=2$, and so on. More formally, it holds for the outcome $o_{flip}(r)$ in round $r$ that

\[
    o_{flip}(r) = 
\begin{cases}
    \text{heads } & \text{if } (r+1) \; {mod}\; 2 = 0\\
    \text{tails } & \text{otherwise}
\end{cases}
\]

\begin{figure}[htbp]
\begin{center}
\includegraphics[width=.8\linewidth]{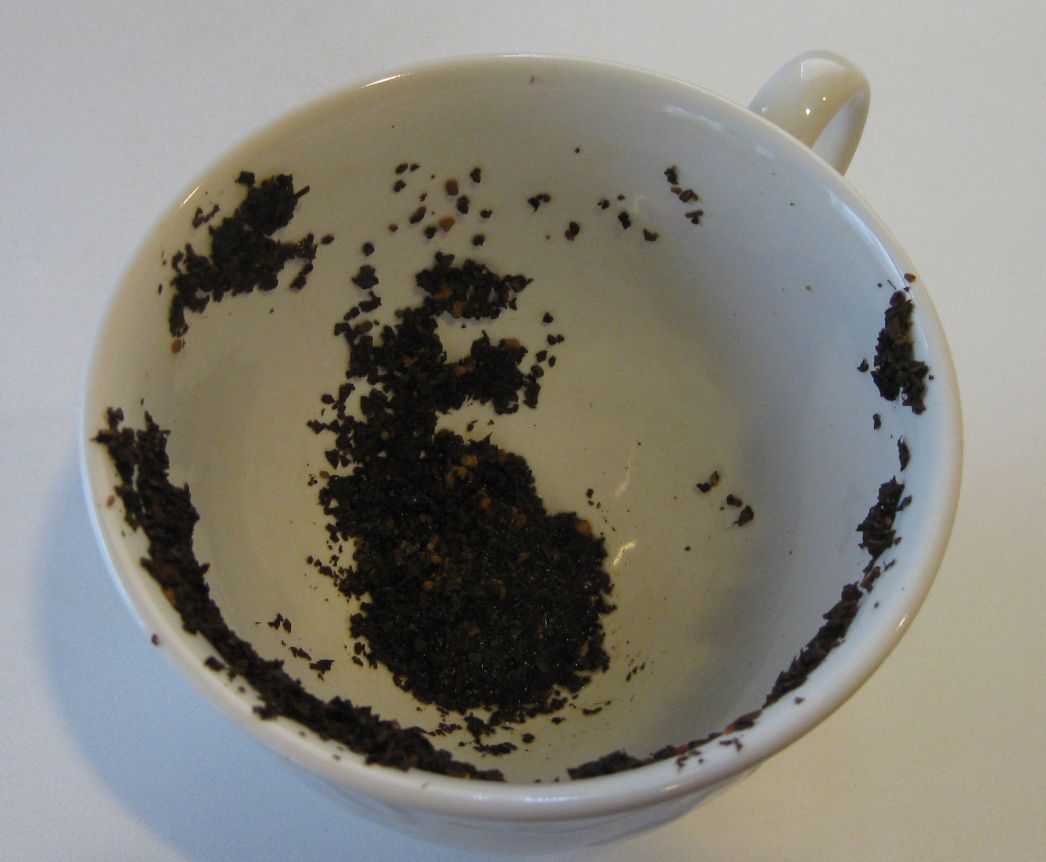}
\caption{A real-world example of a tea-leaves-reading image. \label{fig:cup}}
\end{center}
\end{figure}

\begin{figure*}[htbp]
\begin{center}
\begin{subfigure}[b]{.3\textwidth}
\centering
\includegraphics[width=\textwidth]{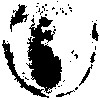}
\caption{The starting image $I_0$ at $r=0$. \label{fig:Tea-leaf-reading-Sim00}}
\end{subfigure}
\begin{subfigure}[b]{.4\textwidth}
\begin{subfigure}[b]{.32\textwidth}
\centering
\includegraphics[width=\linewidth]{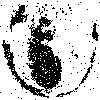}
\caption{$r=1$. \label{fig:Tea-leaf-reading-Sim01}}
\end{subfigure}
\begin{subfigure}[b]{.32\textwidth}
\centering
\includegraphics[width=\linewidth]{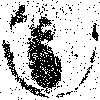}
\caption{$r=2$. \label{fig:Tea-leaf-reading-Sim02}}
\end{subfigure}
\begin{subfigure}[b]{.32\textwidth}
\centering
\includegraphics[width=\linewidth]{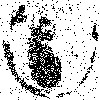}
\caption{$r=3$. \label{fig:Tea-leaf-reading-Sim03}}
\end{subfigure}

\begin{subfigure}[b]{.32\textwidth}
\centering
\includegraphics[width=\linewidth]{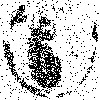}
\caption{$r=4$. \label{fig:Tea-leaf-reading-Sim04}}
\end{subfigure}
\begin{subfigure}[b]{.32\textwidth}
\centering
\includegraphics[width=\linewidth]{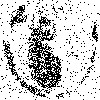}
\caption{$r=5$. \label{fig:Tea-leaf-reading-Sim05}}
\end{subfigure}
\begin{subfigure}[b]{.32\textwidth}
\centering
\includegraphics[width=\linewidth]{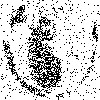}
\caption{$r=6$. \label{fig:Tea-leaf-reading-Sim06}}
\end{subfigure}
\end{subfigure}
\caption{An illustration of the process that generates the images $I_r$. \label{fig:sim}}
\end{center}
\end{figure*}

The images $I_r$ in our high-fidelity simulation are based on a real-world image (Fig.~\ref{fig:cup})\footnote{wikimedia.org; curid=17659832; user: MochaSwirl; Public Domain} of tea-leaves in a cup used by a fortune-telling professional. From this professional image, real-world pixels $p_{tea}$ belonging to tea and real-world pixels $p_{cup}$ belonging to the cup are extracted. This forms the base image $I_0$ with $100 \times 100$ pixels (Fig.~\ref{fig:Tea-leaf-reading-Sim00}). For the sake of simplicity, $p_{tea}$-pixels are shown in black and white is used for $p_{cup}$-pixels . 

The first image $I_1$ is then generated right after the coin was flipped for the first time. So, $I_1$ is linked to producing "head" according to the laws of Tasseology. In general, each image $I_r$ is generated from $I_{r-1}$ by randomly selecting $k_{change}$ coordinates $[x,y]_j : x,y \in \{0,...,99\}$ that are altered as follows 

\[
    I_r[x][y] = 
\begin{cases}
    p_{tea} & \text{if } I_{r-1}[x][y] = p_{cup}\\
    p_{cup} & \text{if } I_{r-1}[x][y] = p_{tea}
\end{cases}
\]


Each $I_r$ is generated right after doing the flip of the coin in round $r$, i.e., it is linked to the outcome $o_{flip}(r)$ of flipping the coin in that round. According to our experience, small to medium values for $k_{change}$ work best. The process is illustrated in Fig.~\ref{fig:sim}. As mentioned, the $p_{cup}$, respectively $p_{tea}$ pixel are shown as white, respectively black pixel in Fig.~\ref{fig:sim} for the sake of simplicity. 

\section{Experiments and Results}
\label{sec:results}  
The high-fidelity simulation presented in Sec.~\ref{sec:sim} is used to generate 500 images. When it is needed by the network, the images are scaled to fit the requirements of PyTorch. The set is split into the first 300 images that are used for training and the following 200 images are used for validation. 

\begin{table}[!htb]
	\centering
	\begin{tabular}{|l|c|c|c|}\hline
	 & YOLOv5 & Resnet-34 & {\bf ours} \\ \hline
accuracy & 53\% & 49\% & {\bf 72\%} \\ \hline
  \end{tabular}
\caption{The accuracy of the three networks. SmileyNet is by far the best. \label{tab:accuracy}}
\end{table}
 
\begin{table}[!htb]
	\centering
	\begin{tabular}{|l|c|c|}\hline
& \multicolumn{2}{c|}{{\bf ours} compared to} \\
 & YOLOv5 & Resnet-34  \\ \hline
accuracy increase & 35.38\% & 46.43\% \\ \hline
data reduction & 66\% & 66\% \\ \hline
  \end{tabular}
\caption{The performance gains of SmileyNet compared to YOLOv5 and Resnet-34. \label{tab:gains}}
\end{table}

Tab.~\ref{tab:accuracy} shows the results for YOLOv5 and Resnet-34. Resnet-34 has only 98 out of 200 cases correctly predicted. YOLOv5 does a bit better and it manages 106 correct predictions on the testing set.

Due to the better performance of YOLOv5 compared to Resnet-34, it is used for a smileyfication to generate SmileyNet version 1. As SmileyNet is expected to have an improved performance, it is trained with less data. Only every fifth image from the data-set is used. Note that this is still an alternating sequence of "heads" and "tails", i.e., it is exactly the same principle as before, but just 100 training images. The remaining 400 images are used for testing.

SmileyNet manages an amazing accuracy of 71.75\%, i.e., 287 out of 400 cases are correctly predicted. This is well above the performance of YOLOv5 and Resnet-34 (Tab.~\ref{tab:accuracy}). Most importantly, it is well above chance and we therefore prove - up to our knowledge - for the very first time that AI can possess psychic powers. Note that it can be shown that the improvement is statistically significant, i.e., it is a thorough and sound result.       

Tab.~\ref{tab:gains} summarizes the performance gains compared to YOLOv5 and Resnet-34. SmileyNet has an accuracy of 71.75\%, i.e., there is an increase by 35.38\% from the 53\% of YOLOv5. The performance gain is in the comparison with Resnet-34 with 46.43\% even higher. SmileyNet requires only 100 training images while YOLOv5 and Resnet-34 had 300 images. So, there is a reduction by 200 images or 66\% in comparison to the other two nets. 

\section{SmileyNet for Lottery Forecasts}
\label{sec:lottery}

The forecasting of throwing a coin can already be a useful psychic power. But the forecast of lotteries has a very straightforward economical potential. Lotteries exist in most countries around the globe with high jackpots for the correctly predicted numbers. It is shown in the following with the example of the German lottery how SmileyNet can be used to this end.

In the German lottery, six numbers $win_1,...,win_6$ are drawn out of a pool with the numbers $1,...,49$. The psychic power of SmileyNet can forecast the flip of a coin, i.e., 1 bit. We therefore simply combine multiple SmileyNets to encode the numbers $1,...,49$. Then, one after the other $win_i$ is predicted. 

More precisely, let $b_j$ denote the forecasting output of SmileyNet ${\cal N}_j$ and $\left\langle b_1,...,b_n \right\rangle_2$ denotes the number $x$ encoded by the bits $b_1,...,b_n$, i.e.,
\[x = \left\langle b_1,...,b_n \right\rangle_2 = \sum_{i=1}^n{b_i \cdot 2^{n-i}}\]

The $i$-th number $win_i$ of a lottery draw is then predicted with ten SmileyNets ${\cal N}_j$ with outputs $b_j$ as follows 
\[win_i = \left\langle b_1,...,b_5 \right\rangle_2 + \left\langle b_6,...,b_9 \right\rangle_2 + \left\langle b_{10}\right\rangle_2\]

As mentioned, the winning numbers $win_1,...,win_6$ are simply generated one after the other. It must of course be ensured that fresh images of tea-leaves from a just emptied cup are provided in every round. Nonetheless, SmileyNet does not have perfect psychic power. It may hence happen in rare occasions that a number $win_{j+k}$ has been predicted before as $win_j$. But this is not a problem. Then, the current $win_{j+k}$ is simply discarded and an other round is added to determine a new $win_{j+k}$.

Given the predictive power of SmileyNet to correctly forecast a bit with $p_{\cal N}=0.7175$, the exact likelihood $p_{win}$ of forecasting the German lottery with the above method can be computed as
\[
\begin{aligned}
p_{win} &= {p_{\cal N}}^5 \cdot {p_{\cal N}}^4 \cdot p_{\cal N} \\ &= 0.03615922 \\ &= 3.615922 \cdot 10^{-2}
\end{aligned}
\]        
This likelihood may look small at first glance, but it is by six orders of magnitude better than just chance with
\[
\begin{aligned}
p_{chance} &= {49 \choose 6}^{-1} \\
&= \left(\frac{49!}{6!(49-6)!}\right)^{-1}\\
&= \frac{1}{13,983,816} \\
&= 7.15112 \cdot 10^{-8}  \\
&\ll p_{win}
\end{aligned}
\]        

This is an important result that demonstrates the huge potential of the novel field of Psychic AI. Note that it is a thorough result as it can be shown that it is statistically significant.

But before we apply these results in practice, we still want to improve the performance of our approach. This holds first and foremost for the likelihood $p_{win}$ of predicting the correct numbers for winning the jackpot:
\begin{itemize}
	\item  One option is simply more training data and more computation power for a larger network to turn SmileyNet into a foundation model for psychic AI. 
\item Furthermore, we expect a transformer to be more accurate given that the drawing of lottery numbers is a weekly event in Germany. This is left for future work.
\end{itemize}
Also, a reduction of the required number of SmileyNets would be nice before applying it in practice as it is a bit inconvenient to drink so much tea. 

\section{Conclusions}
\label{sec:conclusions}
SmileyNet version 1 was introduced. It is based on smileyfication, i.e., the fact that a positive mood can have a positive influence on cognitive functions including classification. It was shown how SmileyNet can be used to predict the lottery by reading tea-leaves. First, the flipping of a coin was predicted with a very good accuracy, namely 72\%. State of the art networks like YOLOv5 and Resnet-34 perform much worse. Second, it was shown how multiple SmileyNets can be combined for lottery forecasts. In addition to the potential practical implications, it is a major breakthrough to show that AI can have psychic powers.  

\section{Acknowledgments}
We thank our anonymous reviewers for their feedback. Special thanks go to reviewer 2, who pointed out that the drawing of lottery numbers tends to be a weekly or at least regular event, and that the use of a transformer should be beneficial. 

\newpage
\onecolumn 
\section*{Epilog}
This is a satirical accumulation of misconceptions, mistakes, and flawed reasoning I have encountered in recent times as a reviewer and sometimes even as a reader of published papers. I hope it is entertaining and useful in the context of the education of BSc, MSc, and PhD students in Machine Learning, Artificial Intelligence, and Cognitive Science. 

For an easy start, here a few questions to reflect upon. First and foremost, there is of course the following puzzle: 
\begin{itemize}
	\item {\bf Where does the "magic" hide that is presented here? How can SmileyNet "predict" the flip of a coin better than chance? }
\end{itemize}
As a small hint: there is a single conceptual mistake that explains it. 

There are also many other bugs, respectively misconceptions and flawed reasoning in this paper. Here some further food for thought, to hopefully teach a good lesson by setting a bad example:   
\begin{itemize}
	\item Can AI beat natural laws? What are the pros and cons of a detailed pre-analysis of the possible information content of the data?
	\item How much academic novelty is in SmileyNet? How much does it differ from the net it is build upon? 
	\item How do we prevent {\em smileyfication} in our own work? How can we properly show that an "extension" or "modification" of an existing approach leads to a substantial improvement and study the related trade-offs?  
	\item What is a proper assessment of the state of the art? How do we find, understand, and discuss related work?
	\item What is a proper statistical performance analysis? What is statistical significance and how should it be interpreted?
	\item When and how do I expect my own work to be of relevance? Why do I publish it? 
	\item What are Artificial Intelligence and Machine Learning? Which sub-areas within these disciplines exist? What are commonly used methodologies within the sub-areas? 
		\item What is the role of training data in supervised machine learning? What are the pros and cons of simulation?
	\item What is Open Science? What are its principles? What does it imply for the quality of the underlying work? What are the pros and cons if others, respectively oneself adheres to it? 
	\item What is a bio-inspired design? How can insights from Cognitive Science be used and which pitfalls exist?
	\item What are the pros and cons of bibliometrics? How can they help us to identify relevant work? And where are their constraints and limitations?
\end{itemize}
This list is a bit concentrated on the "big questions". It is of course also possible to just concentrate on the (many) methodological mistakes in the SmileyNet paper. So, the above list is neither meant to be exhaustive nor to be without dispensable aspects depending on the context. It is just a possible starting point for using the SmileyNet in the class-room, or to just meditate about it on your own, or... 

Enjoy and have fun...

\end{document}